\documentclass{article} 
\usepackage{iclr2027_conference,times}

\usepackage{amsmath,amsfonts,bm}

\def\eqref#1{equation~\ref{#1}}

\def\1{\bm{1}}

\DeclareMathAlphabet{\mathsfit}{\encodingdefault}{\sfdefault}{m}{sl}
\SetMathAlphabet{\mathsfit}{bold}{\encodingdefault}{\sfdefault}{bx}{n}

\usepackage{url}
\usepackage{booktabs}       
\usepackage{amsfonts}       
\usepackage{nicefrac}       
\usepackage{microtype}      
\usepackage{xcolor}         
\usepackage[pagebackref,breaklinks,colorlinks]{hyperref} 
\hypersetup{
 colorlinks=true,
 linkcolor=red,
 citecolor=teal,
 filecolor=magenta, 
 urlcolor=magenta,
 }
\usepackage{enumitem}
\usepackage{amsmath}
\usepackage{multirow}
\usepackage{threeparttable}
\usepackage{graphicx}
\usepackage{subcaption}

\usepackage{algorithm}
\usepackage{algorithmic}
\usepackage{float}

\title{Guided Trajectory Optimization with Sparse Scaling for Test-Time Diffusion}

\author{%
  \centerline{Gang Dai$^{1}$, Yining Huang$^2$, Yiming Xia$^2$, Guohao Chen$^3$, Shuaicheng Niu$^{3*}$} \\
  \centerline{$^{1}$Guangdong University of Technology,} \\
  \centerline{$^{2}$South China University of Technology,
  $^{3}$Nanyang Technological University} \\
}

\iclrfinalcopy 
\begin{document}

\maketitle

\begin{abstract}
Test-Time Scaling (TTS) paradigm offers a promising perspective for enhancing the generation performance of diffusion models. However, current solutions largely restrict their search to predefined noise candidates or suffer from inflexible exploration across the denoising trajectory. To bridge this gap, we propose RTS, a novel Reward-guided Trajectory Scaling method to fully unlock the generative potential of diffusion models. Unlike existing methods, RTS facilitates the synthesis of refined, high-fidelity images via two core innovations: 1) a coarse-to-fine noise optimization mechanism that exploits historical search experience to actively steer the exploration toward high-reward regions and 2) a unified sparse test-time scaling framework featuring PCA-driven curvature analysis, which eliminates temporal redundancy by flexiblely allocating compute to a sparse set of key timesteps that represent critical shifts in the denoising direction. Extensive experiments across SD v3, FLUX, and Qwen-Image architectures demonstrate that RTS outperforms baselines, improving the GenEval score by $20.7\%$, $15.6\%$, and $12.2\%$, respectively. Notably, empirical findings indicate that these key points primarily cluster in the mid-stage of the trajectory, distinct from the structure-sensitive early phases and the late attribute refinement phases.
\end{abstract}


\section{Introduction}
Generative models have revolutionized domains such as vision~\citep{rombach2022high}, language~\citep{achiam2023gpt}, and biology~\citep{watson2023novo}, demonstrating exceptional capabilities to model complex data distributions. In the field of visual synthesis, diffusion models~\citep{esser2024scaling} have become a powerful architecture for generating high-quality images. Key to their success is the flexibility to boost performance through scaling model size~\citep{hoffmann2022training}, fine-tuning with labeled data~\citep{clark2023directly}.
However, these training-based methods are increasingly constrained by their reliance on exorbitant computational costs and extensive labeled data.

\begin{figure}[t]
\centering
\includegraphics[width=0.85\textwidth]{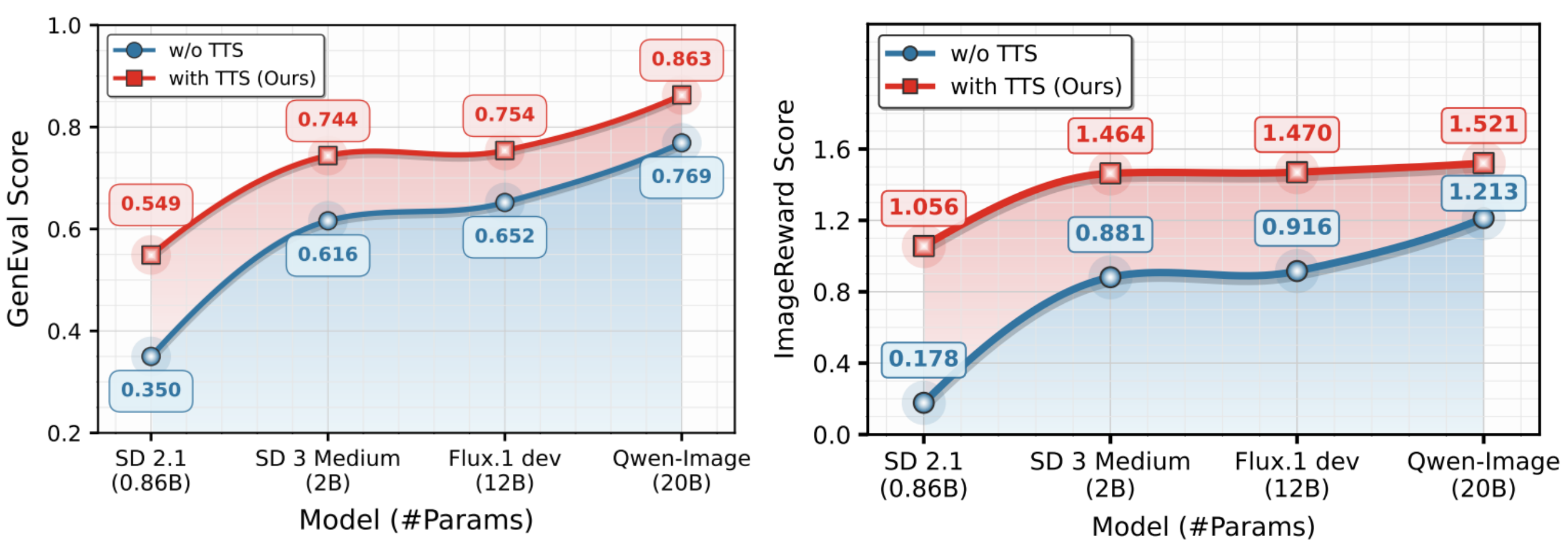} 
\caption{Performance evaluation of different generation models on \textbf{left}: GenEval and \textbf{right}: ImageReward benchmarks, respectively. TTS paradigm empowers lightweight models to achieve performance superior to that of multi-fold larger models. For example, the 2B SD 3 Medium equipped with TTS significantly outperforms the 12B Flux.1 dev on GenEval (0.744 vs. 0.652), achieving results highly competitive with the 20B Qwen-Image (0.769).}
\label{fig:sub1}
\vspace{-0.2in}
\end{figure}

Recently, a novel paradigm called Test-Time Scaling (TTS)~\citep{brown2024large,snell2024scaling,wu2408empirical} has garnered widespread attention. Unlike training-based methods, TTS focuses on fully leveraging a model's inherent capacity while requiring minimal computational resources during the inference stage. This new paradigm offers a fresh perspective for enhancing the performance of diffusion models (cf. Figure~\ref{fig:sub1}). Building on this, we aim to explore the seamless integration of diffusion models with TTS techniques to boost image generation performance.

To unlock the potential of diffusion models, early TTS methods, such as Best-of-N and Zeroth-order search~\citep{ma2025inference} run multiple initial noises through a diffusion solver to generate candidates and select the best based on reward scores. While straightforward, these methods neglect the impact of intermediate noises, thereby limiting the models' inherent generative capacity. In response, particle-based search strategies~\citep{singhal2025general} treat noisy samples as particles and resample them at intermediate time steps based on weight factors, enabling exploration across the entire denoising trajectory. However, these approaches are limited to selecting trajectories starting from predefined noise candidates, lacking the flexibility to dynamically generate new particles beyond the initial set. In summary, existing approaches exhibit two major limitations: 1) Relying heavily on predefined noise candidates, which restricts the ability to explore potential high-reward regions. 2) Failing to simultaneously consider initial and intermediate noises results in incomplete search coverage.

In this paper, we propose \textbf{RTS}, a novel \textbf{R}eward-guided \textbf{T}rajectory \textbf{S}caling method to unleash the generative potential of diffusion models. RTS comprises two components: 1) A reward-guided noise optimization strategy to identify high-reward noise instances. 2) A unified test-time scaling framework that incorporates a sparse optimization strategy, accounting for both initial and intermediate noises.

Finding optimal noise instances in the vast noise space is challenging. Previous methods~\citep{ma2025inference,ramesh2025test} rely on stochastic sampling, which is highly inefficient due to the lack of specific guidance. To address this, we leverage historical search experience to accelerate the exploration process. This is achieved through a coarse-to-fine alternating mechanism, which alternates between gaining directional insights via coarse-level estimation and utilizing them for targeted, fine-grained exploration. During the coarse-level estimation, we apply local perturbations to the base noise to generate neighboring samples. By analyzing their performance relative to the base noise, we estimate a surrogate gradient direction as our refined search direction. In the fine-grained exploration, this refined direction guides the generation of new neighboring samples, adaptively steering the search toward high-quality regions. Thus, this iterative process significantly enhances generation quality by effectively exploiting historical efforts.


To jointly optimize initial and intermediate noises, we regard the iterative transformation from Gaussian noise to a clean image as a complete denoising trajectory. From this perspective, we introduce a unified test-time scaling framework (cf. Figure~\ref{fig1:method}) to boost diffusion models. Specifically, for the initial noise, we employ multiple coarse-to-fine alternating search cycles to identify an optimal starting point. We then refine the intermediate noise states along the denoising path using a single alternating cycle. To effectively guide this refinement, we introduce a PCA-driven curvature analysis that explicitly bypasses temporal redundancy. By pinpointing critical state transitions at high-curvature points where the trajectory drastically shifts, we can precisely allocate resources to steer the generation where intervention matters. Our main contributions are summarized as follows:

\begin{itemize}[leftmargin=*]
    \item RTS introduces a sparse test-time scaling framework that enhances diffusion model performance by flexibly allocating compute along the denoising trajectory.
    \item RTS proposes a novel reward-guided noise optimization strategy for diffusion models to proactively explore high-reward regions within the noise space.
    \item Extensive quantitative and qualitative experiments on GenEval and T2I-CompBench datasets demonstrate that our RTS achieves state-of-the-art performance in terms of generation quality.
\end{itemize}

\section{Related Work}
\label{sec:relatedwork}

\textbf{Test-time scaling for diffusion.} 
While Test-time scaling (TTS) is well-established in the realm of large language models~\citep{brown2024large,snell2024scaling,wu2408empirical}, however, its integration with diffusion models remains an underexplored area of research. An early TTS attempt for diffusion models leverages more denoising steps, which is later found to cause diminishing returns and error accumulation~\citep{xu2023restart}. Beyond diffusion steps, recent studies~\citep{ma2025inference,singhal2025general,ramesh2025test} optimize noise representations to achieve more effective scaling. Specifically, several sampling-based search algorithms~\citep{ma2025inference} such as Best-of-N (BoN) search and Zeroth-Order (ZO) search are proposed. Specifically, BoN sampling generates multiple independent noise vectors and selects the candidate achieving the highest reward after full denoising. Meanwhile, ZO iteratively refine the initial noise based on reward feedback. However, both algorithms are conducted on ODE-based diffusion models, which possess a deterministic mapping from the initial noise $\mathbf{x}_t$ to the final sample $\mathbf{x}_0$. Thus, they focus exclusively on the randomness of the initial noise.

To address above limitations, recently, some studies propose particle sampling, such as the Feynman-Kac (FK) formalism \citep{singhal2025general}. It focuses on scaling the denoising trajectory by resampling noise representations (\emph{i.e.,} particles) at intermediate steps based on scores computed using reward functions, thereby effectively steering the generation process through state space. However, FK primarily filters existing candidates rather than actively exploring new regions of the state space, which restricts state-space diversity. Meanwhile, NTS~\citep{ramesh2025test} relies on a greedy sampling strategy that requires drawing multiple candidate latents at every denoising step renders it highly inefficient computationally.

In contrast, our RTS proposes a sparse test-time framework coupled with a reward-guided noise exploration strategy, fully unleashing the generative potential of the diffusion model. We discuss additional related work on \textbf{Zeroth-order search} and \textbf{Diffusion models} in Appendix~\ref{sec:more_related}.

\section{Problem Statement and Preliminaries}

\textbf{Problem statement.} Given a text prompt $c$ and a pre-trained diffusion model $\mathbf{D}_\theta$, our goal is to synthesize an optimal sample $\mathbf{x}_0^*$ that maximizes the reward paradigm $\mathcal{R}$ during inference. In the denoising generation process, the transition from initial Gaussian noise to the final image is governed by a sequence of stochastic noise components, collectively forming the denoising trajectory $\mathcal{Z}$. Thus, the pursuit of optimal sample can be reformulated as searching for optimal noise trajectory $\mathcal{Z}^*$, formally expressed as:
\begin{equation}
     \mathcal{F}(\mathbf{D}_\theta ,\mathbf{c},\mathcal{R},\mathcal{Z}  ) \rightarrow\mathbf{x}_0^*,
\end{equation}
where $\mathcal{F}(\cdot)$ denotes the test-time scaling objective. The trajectory $\mathcal{Z} = \{\mathbf{z}_L, \mathbf{z}_{path}\}$ includes the initial Gaussian latent $\mathbf{z}_L \in \mathbb{R}^d$ and $\mathbf{z}_{path} = \{\mathbf{z}_l\}_{l=1}^{L-1}$ represents the sequence of stochastic noise injected at subsequent steps.

The main challenge lies in the high dimensionality of the noise trajectory $\mathcal{Z} \in \mathbb{R}^{L \times d}$, leading to exponential expansion of the search space. To address this, we employ sparse optimization to identify key denoising steps, followed by a reward-guided strategy to flexibly explore high-reward noise regions
at each key step. This enables effective navigation within complex noise spaces.

\textbf{Applicability to ODE and SDE solvers.} The denoising process is typically characterized by either Ordinary Differential Equations (ODEs) or Stochastic Differential Equations (SDEs). ODE-based solvers provide a deterministic mapping from the initial latent $\mathbf{z}_L$, whereas SDE-based solvers introduce additional stochasticity via intermediate noise $\mathbf{z}_{path}$. For a unified solution, we optimize the initial noise to enhance the starting state for both ODE and SDE solvers. Beyond this, we conduct intermediate trajectory searches exclusively for SDEs, as deterministic ODE solvers cannot naturally generate multiple diverse candidate paths midway through the trajectory. By adapting to these distinct generative dynamics, our RTS ensures robust performance gains across various solver paradigms.

\begin{figure*}[t]
\centering
\includegraphics[width=0.95\textwidth]{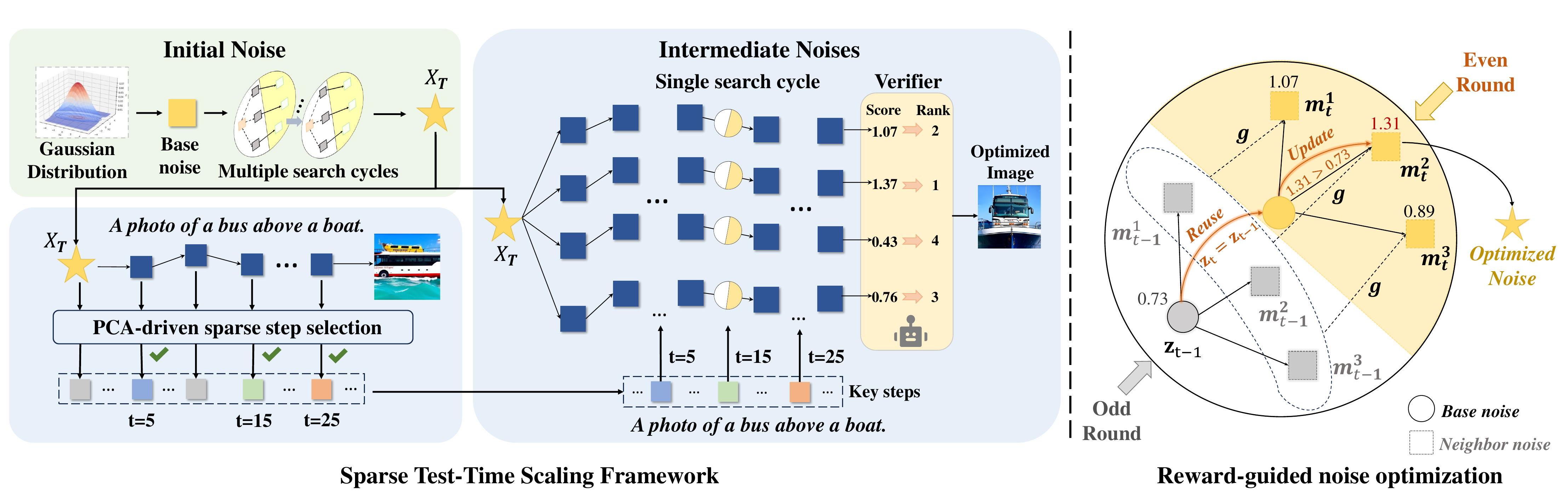} 

\caption{Overview of our method. \textbf{Left:} The sparse test-time scaling framework bridges the optimization of initial and intermediate noises (cf. Section~\ref{pca}). Specifically, an optimal initial noise is first identified to generate a base denoising trajectory via a pre-trained diffusion model. PCA-driven analysis is then applied to this specific trajectory to select sparse key timesteps for intermediate noise refinement. \textbf{Right:} The reward-guided strategy iteratively guides both initial and intermediate search processes toward high-reward regions via coarse-to-fine alternating rounds (cf. Section~\ref{active}).}
\label{fig1:method}
\vspace{-0.2in}
\end{figure*}

\section{Methodology}
\label{sec:method}

To enhance the generation quality of diffusion models, we seek to search for superior noise instances in the denoising trajectory. A naive approach involves random sampling across the entire noise path to find the best candidates. However, random sampling in such a high-dimensional space is highly inefficient, and a full-path search is computationally prohibitive. To address this, we propose RTS, a novel test-time scaling method for effective noise optimization. As shown in Figure~\ref{fig1:method}, our RTS consists of two main compoments: a sparse optimization framework (cf. Section~\ref{pca}) to narrow the search space and alleviate computational overhead, and a reward-guided strategy (cf. Section~\ref{active}) to provide gradient-driven exploration during the initial and intermediate stages.



\subsection{Sparse Test-Time Scaling Framework}
\label{pca}
To fully unleash the generative potential of diffusion models, we propose a new Test-Time Scaling framework to optimize both the initial and intermediate noises. However, optimizing the entire denoising path is computationally intensive. To flexibly allocate resources, we propose a sparse optimization strategy using PCA-based dimensionality reduction (cf. Equation~(\ref{svd})) to analyze each denoising step and identify a sparse set of key inflection points. Based on this sparse scaling, our framework optimizes the entire noise trajectory with low computational cost (cf. Figure~\ref{fig1:method}). Specifically, for the initial noise, we apply multiple coarse-to-fine alternating search cycles to identify an optimal starting point. For each point from the sparse set, we execute a single coarse-to-fine cycle to obtain an optimal intermediate noise (cf. \textbf{Appendix}~\ref{intermediate} for more optimization details.). By jointly optimizing these noise components, our framework achieves comprehensive coverage of the entire denoising space, ensuring robust performance while drastically reducing the redundant computation.

\begin{figure*}[t]
\centering
\includegraphics[width=0.8\textwidth]{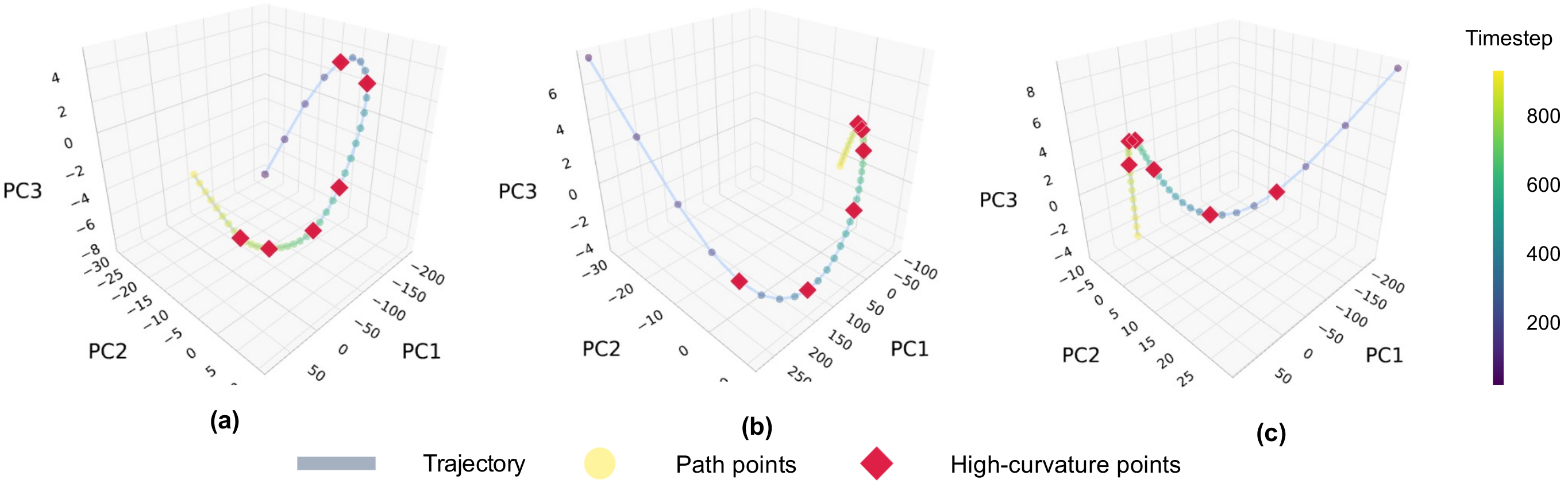} 
\caption{Three visual examples illustrating PCA-based key-step selection along different denoising trajectories. The red diamonds highlights the identified high-curvature points.}
\label{pca_points}
\vspace{-0.2in}
\end{figure*}

\textbf{PCA-driven sparse step selection.} We employ a PCA-driven analysis to identify key steps within the denoising trajectory for flexible resource allocation. This sparse selection builds on the inherent temporal redundancy~\citep{Liu0W0QZZY025, ChuWFZ25} of the denoising process. Smooth trajectory segments indicate a consistent generation direction, making scaling interventions largely redundant. Conversely, high-curvature points mark timesteps where the trajectory direction undergoes drastic corrections. Since these directional shifts largely dictate the final generated outcome, targeting these specific moments for intervention effectively improves generation quality. To this end, we collect the high-dimensional latent vectors $\{\mathbf{x}_l \in \mathbb{R}^d \mid l = 1, \ldots, L\}$ from the complete denoising process, construct a centered data matrix $\mathbf{X} \in \mathbb{R}^{L \times d}$, and perform Principal Component Analysis (PCA) via Singular Value Decomposition:
\begin{equation}
\label{svd}
\mathbf{X} = \mathbf{U} \mathbf{\Sigma} \mathbf{V}^T.
\end{equation}
Our preliminary analysis reveals that the first three principal components (columns of $\mathbf{V}$) capture the most significant variance of the entire denoising trajectory (cf. Appendix~\ref{pca-ablation} for detailed discussions and ablation studies). To strike an optimal balance between trajectory representation and information redundancy, we project the high-dimensional latents into a 3D subspace: $\mathbf{p}_l = \mathbf{X}_l \mathbf{V}_{:3}$, yielding the trajectory points $\{\mathbf{p}_l \in \mathbb{R}^3\}$ (cf. Figure~\ref{pca_points}).

To systematically pinpoint the critical transitions, we quantify the local curvature $f(\mathbf{p}_l)$ of the path at each projected point (cf. Appendix~\ref{sec:loca}). By ranking these curvature values, we define a sparse set of key inflection points as $\mathcal{P}_{\text{key}} = \text{Top-}k \big( \{ f(\mathbf{p}_l) \}_{l=1}^L \big)$. Consequently, we focus our optimization efforts on the initial noise and these sparse inflection points, rather than conducting a uniform search across all steps. Crucially, because this curvature analysis is strictly driven by the dynamics of the specific denoising trajectory (cf. Figure~\ref{pca_points}), the selected inflection points adaptively vary across different generation contexts (\emph{e.g.,} prompts and model backbones). Furthermore, empirical validations confirm that these keypoints are robust to optimization-induced trajectory alterations, allowing us to efficiently share them across all search rounds without recalculation (cf. Appendix~\ref{roubutness-key}).

\subsection{Reward-Guided Noise Optimization}
\label{active}
To effectively navigate the search process toward high-reward regions, we propose a reward-guided noise optimization strategy centered on a coarse-to-fine alternating mechanism, as illustrated in Figure~\ref{fig1:method} and \textbf{Algorithm}~\ref{alg:adaptive_zo}. To identify better directions for improvement, we generate a set of neighbor candidates $\mathcal{M} = \{\mathbf{m}^1, \ldots, \mathbf{m}^N\}$ by introducing random perturbations to a base noise $s$. We then use the collective reward feedback from these candidates to estimate the direction of steepest ascent, guiding the sampling of higher-reward candidates. In practice, the search process alternates between: 1) generating multiple neighbor candidates to facilitate gradient estimation, and 2) applying the estimated gradient to guide refined neighbor noise generation. The details of the alternating mechanism, neighborhood construction, and gradient estimation are provided below.


\textbf{Coarse-to-fine alternating mechanism.} To effectively navigate the reward landscape, we structure the optimization process into alternating cycles of coarse-grained and fine-grained search rounds. Specifically, a coarse-grained search aims to stochastically explore the surrounding region to estimate a surrogate gradient. In the initial round ($t=1$), the base noise $\mathbf{s}_1$ is randomly initialized from a Gaussian distribution, $\mathbf{s}_1 \sim \mathcal{N}(0, \mathbf{I})$. In any subsequent coarse-grained round ($t \in \{3, 5, \dots\}$), the base noise $\mathbf{s}_t$ is updated using a greedy relocation rule. This is based on the best-performing vector $\mathbf{m}_{t-1}^*$ from all neighbor candidates $\mathcal{M}_{t-1}$ evaluated in the preceding round $t-1$:
\begin{equation}
\mathbf{s}_{t} = 
\begin{cases}
\mathbf{m}_{t-1}^* & \text{if } R(\mathbf{m}_{t-1}^*) > R(\mathbf{s}_{t-1}), \\
\mathcal{N}(0, \mathbf{I}) & \text{otherwise}.
\end{cases}
\end{equation}
Under this rule, $\mathbf{m}_{t-1}^*$ replaces the previous base $\mathbf{s}_{t-1}$ if it achieves a higher reward $R$. Otherwise, a new random base is sampled from a Gaussian distribution to avoid getting stuck in local minima. After updating the base noise, neighbor noise candidates $\mathcal{M}_{t} = \{\mathbf{m}_t^1, \ldots, \mathbf{m}_t^N\}$ are constructed via random spherical interpolation (cf. Equation~(\ref{rand_qph})), and the proxy gradient $\mathbf{g}_t$ is estimated based on differential signals of these neighbor candidates (cf. Equation~(\ref{est_grad})).

Conversely, a fine-grained search is conducted for targeted refinement guided by the estimated gradient. During a fine-grained round $t \in \{2, 4, \dots\}$, we reuse the base noise from previous round $t-1$ and generate new neighbors by integrating the estimated gradient from round $t-1$ with a perturbation component (cf. Equation~(\ref{guided_qph})). This fusion balances historical gradients and stochasticity, ensuring robust and stable progress. The gradient-informed exploration directs sampling toward regions anticipated to yield higher rewards.

In summary, the iterative execution of multiple coarse-to-fine alternating rounds accelerates the search process. This mechanism marks a pivotal shift from stochastic exploration to actively targeting high-reward regions, improving the effectiveness of the optimization process.

\textbf{Spherical neighborhood construction.} The goal of neighborhood construction is to generate noise neighbors through controlled perturbations, thus exploring new candidates within local regions. To achieve this, it is crucial that perturbations preserve the high-dimensional norm of the noise vectors, as generative models are sensitive to the magnitude of latent vectors. This ensures that noise candidates remain within the learned distribution rather than drifting into low-probability regions. We use spherical interpolation sampling to maintain the norm of the noise vectors, preventing degradation in generative quality or the production of out-of-distribution artifacts. Formally, let
$\mathbf{u} = \frac{\mathbf{z}}{\|\mathbf{z}\|}$
 be the unit direction of base vector $\mathbf{z}$ and $r={\|\mathbf{z}\|}$ be the radius. 
 
 In coarse-grained search rounds, we draw $N$ random Gaussian noises $\{\mathbf{w}_i\}_{i=1}^N$ and project them onto the orthogonal complement of $\mathbf{u}$ to ensure they are strictly tangential:
\begin{equation}
\label{rand_qph}
\begin{gathered}
\mathbf{w}_i \sim \mathcal{N}(0, \mathbf{I}), \quad
\mathbf{w}_i^{\perp} = \mathbf{w}_i - (\mathbf{w}_i \mathbf{u}^{\top}) \mathbf{u}, \quad
\hat{\mathbf{w}}_i = \frac{\mathbf{w}_i^{\perp}}{\|\mathbf{w}_i^{\perp}\|}, \\
\mathbf{m}^i = r \cdot \left( \tau \cdot \mathbf{u} + \sqrt{1 - \tau^2} \cdot \hat{\mathbf{w}}_i \right),
\end{gathered}
\end{equation}
where $\tau \in (0,1)$ controls the angular deviation from the base direction, $\mathbf{m}^i$ represents a generated neighbor noise. In fine-grained search rounds, we utilize the estimated surrogate gradient $\mathbf{g}$ to guide the neighbor generation. We update the neighborhood noises by refining the perturbation term $\hat{\mathbf{w}_i}$:
\begin{equation}
\label{guided_qph}
\begin{gathered}
\mathbf{g}^\perp = \mathbf{g} - (\mathbf{g} \mathbf{u}^\top) \mathbf{u}, \hat{\mathbf{w}}_i = (1 - \alpha) \cdot \hat{\mathbf{w}}_i' + \alpha \cdot \frac{\mathbf{g}^\perp}{\|\mathbf{g}^\perp\|}, 
 \\
\mathbf{m}^i = r \cdot \left( \tau \cdot \mathbf{u} + \sqrt{1 - \tau^2} \cdot \frac{{\hat{\mathbf{w}}_i}}{\|\hat{\mathbf{w}}_i\|} \right).
\end{gathered}
\end{equation}
Here, $\hat{\mathbf{w}}_i'$ represents the random perturbation from the previous round, while $\alpha \in [0,1]$ is an interpolation factor that controls the degree of historical exploration momentum retained in the current update. Further discussions on this interpolation choice are provided in the Appendix~\ref{choice}.

\textbf{Surrogate gradient estimation.} While gradient-based guidance offers an effective solution for exploration in high-dimensional space, standard backpropagation is limited by the need for differentiable rewards and significant computational latency. Consequently, steering diffusion generation with arbitrary rewards during inference time remains a challenge. To overcome this, we estimate a surrogate gradient $\mathbf{g}$ that synthesizes robust directional guidance without requiring backpropagation.
Specifically, given a base latent $\mathbf{z}$ and its reward score $R(\mathbf{z})$ evaluated by a verifier, we generate $N$ neighbors $\mathcal{M} = \{\mathbf{m}^1, \ldots, \mathbf{m}^N\}$ and unit orthogonal component $\hat{\mathbf{w}}_i$ using Equation~(\ref{rand_qph}). After evaluating their rewards $\{R(\mathbf{m}_i)\}_{i=1}^N$, the surrogate gradient is computed as:
\begin{equation}
\label{est_grad}
\mathbf{g}= \sum_{i=1}^N \left( \frac{R(\mathbf{m}_i)-R(\mathbf{z})}{\sum_{i=1}^NR(\mathbf{m}_i)+R(\mathbf{z})} \right) \cdot \hat{\mathbf{w}}_i.
\end{equation}
By aggregating relative rewards from neighboring samples, this formulation estimates a zero-order gradient $\mathbf{g}$ to synthesize robust directional guidance within the noise space. A detailed justification for this formulation is provided in Appendix~\ref{Derivation}.

\section{Experiment}
\label{sec:exper}

\textbf{Evaluation dataset.} To evaluate our RTS on text-to-image generation task, we use the widely adopted GenEval~\citep{singhal2025general,esser2024scaling,xie2024show} and T2I-CompBench~\citep{huang2023t2i,ma2025inference,zhang2025compass} benchmark datasets. GenEval (553 prompts) focuses on fine-grained object attributes (\emph{e.g.}, object co-occurrence, position, count, and color) and spatial reasoning. T2I-CompBench (1800 prompts) evaluate attribute binding, object relationships, and complex compositions. For each benchmark, we generate two images per prompt.

\textbf{Implementation details.} In our experiments, we use the official Stable Diffusion v3 (SD3), FLUX-1.dev, and Qwen-Image as backbone models, with FLUX serving as the primary one for extended analysis. For the reward-guided noise optimization strategy (cf. \textbf{Algorithm}~\ref{alg:adaptive_zo}), we set neighbor count $N=3$, weight $\alpha=0.7$, and balance factor $\tau=0.9$ by conducting grid search. When exploring the initial noise, we set the search round $T$ to $4$ and perform $L=50$ denoising steps to predict the clean image to receive reward. For intermediate noises, we set $T=2$ and $L=1$ to expedite the search process. Through ablation analysis (cf. Table~\ref{tab:my-table}), we select ImageReward~\citep{xu2023imagereward} as the reward model $R$.We empirically set the number of key steps $k$ to 6 to balance performance and efficiency (cf. Appendix Table~\ref{tab:key_steps_ablation}). We provide more implementation details in Appendix~\ref{hyperparameter}.

\textbf{Baselines.} We compare our method against state-of-the-art noise optimization approaches, including Best-of-N (BoN)~\citep{ma2025inference} and Zeroth-Order (ZO)~\citep{ma2025inference}, Feynman-Kac (FK)~\citep{singhal2025general}, and Noise Trajectory Search (NTS)~\citep{ramesh2025test}.

\begin{table*}[t]
\centering
\caption{Quantitative comparison of our method with SOTA methods on text-to-image generation in GenEval benchmark. We select SD v3 and Flux-1.dev, and the advanced Qwen-Image as backbones for detailed investigation. To ensure an apples-to-apples comparison, we configure all TTS methods to utilize the exact same ImageReward verifier. The ``Held-in metric'' refers to the evaluation metric used as the reward model during optimization, whereas ``Held-out metrics'' are strictly unseen and never used to guide the search process. Human-preference results are provided in Appendix Table~\ref{tab:user_study}.}%

\label{tab1}
\resizebox{1.0\linewidth}{!}{
\begin{tabular}{lcccccccc}
\hline
\multirow{2}{*}{TTS Method} & \multirow{2}{*}{Base Model} & \multirow{2}{*}{\#Params} & Held-in Metric & \multicolumn{3}{c}{Held-out Metrics} & \multirow{2}{*}{NFEs} & \multirow{2}{*}{\shortstack{Runtime (s)}} \\
\cmidrule(r){4-4} \cmidrule(l){5-7} 
& & & ImgReward $\uparrow$ & GenEval $\uparrow$ & NIQE $\downarrow$ & CLIP $\uparrow$ & & \\ 
\hline
\multirow{4}{*}{TTS Free} 
& SD v2.1    & 0.86B &   0.1779        &     0.3307      & 4.4351          & 0.2701          & 28   & 1.5 \\
& SD v3      & 2B    &   0.8814       &     0.6165     & 4.3724          & 0.2631          & 28   & 1.7 \\
& Flux-1.dev & 12B   &   0.9159      &   0.6516     & 4.2170          & 0.2727          & 50   & 9.8 \\
& Qwen-Image & 20B   &  \textbf{1.2131} & \textbf{0.7692} & \textbf{4.0349} & \textbf{0.2935} & 50   & 11.2s \\ 
\hline

BoN~\citep{ma2025inference} 
& \multirow{4}{*}{SD v3} & \multirow{4}{*}{2B}  
&  1.3176         &     0.7054      & 4.1951          & 0.2867          & 560  & 34.1 \\
ZO~\citep{ma2025inference} 
&                         &                       
&       1.3842    &    0.7016       & 4.1742          & 0.2896          & 560  & 35.9 \\
FK~\citep{singhal2025general} 
&                         &                       
&  1.3465         &    0.7126       & 4.1810          & 0.2904          & 560  & 44.2 \\
Ours 
&                         &                       
&  \textbf{1.4636} & \textbf{0.7438} & \textbf{4.0734} & \textbf{0.2936} & 560  & 38.7 \\ 
\hline

BoN~\citep{ma2025inference} 
& \multirow{4}{*}{Flux-1.dev} & \multirow{4}{*}{12B} 
&     1.3192      &       0.7101    & 4.0454          & 0.2878          & 1000 & 196.5 \\
ZO~\citep{ma2025inference} 
&                             &                         
&   1.3355        &   0.7168        & 3.9855          & 0.2791          & 1000 & 200.4 \\
FK~\citep{singhal2025general} 
&                             &                         
&     1.3679      &   0.7206        & 3.9587          & 0.2889          & 1000 & 239.1 \\
Ours 
&                             &                         
&  \textbf{1.4703} & \textbf{0.7535} & \textbf{3.8506} & \textbf{0.2964} & 1000 & 217.7 \\ 
\hline

FK~\citep{singhal2025general} 
& \multirow{2}{*}{Qwen-Image} & \multirow{2}{*}{20B} 
&      1.4471     &     0.8109      & 3.9238          & 0.2974          & 1000 & 292.3 \\
Ours 
&                             &                      
&  \textbf{1.5214} & \textbf{0.8634} & \textbf{3.9147} & \textbf{0.3061} & 1000 & 260.4 \\ 
\hline
\end{tabular}
\vspace{-0.2in}
}

\end{table*}

\textbf{Metrics.} 
In addition to the standard GenEval score~\citep{ghosh2023geneval}, we assess semantic alignment using CLIP score~\citep{hessel2021clipscore} and ImageReward~\citep{xu2023imagereward} for predicted human ratings. Perceptual quality is quantified via NIQE~\citep{mittal2012making}, and computational cost is measured by the Number of Function Evaluations (NFEs). We also conduct
user studies to quantify the subjective quality of the generated images in Appendix~\ref{sec:user}. For T2I-CompBench, following~\citep{huang2023t2i,singhal2025general}, we employ BLIP-VQA for attribute binding, UniDet for spatial relationships, and a weighted average of BLIP, UniDet, and CLIP scores for complex compositions.

\begin{figure*}[t]
\centering
\vspace{-0.15in}
\includegraphics[width=0.9\textwidth]{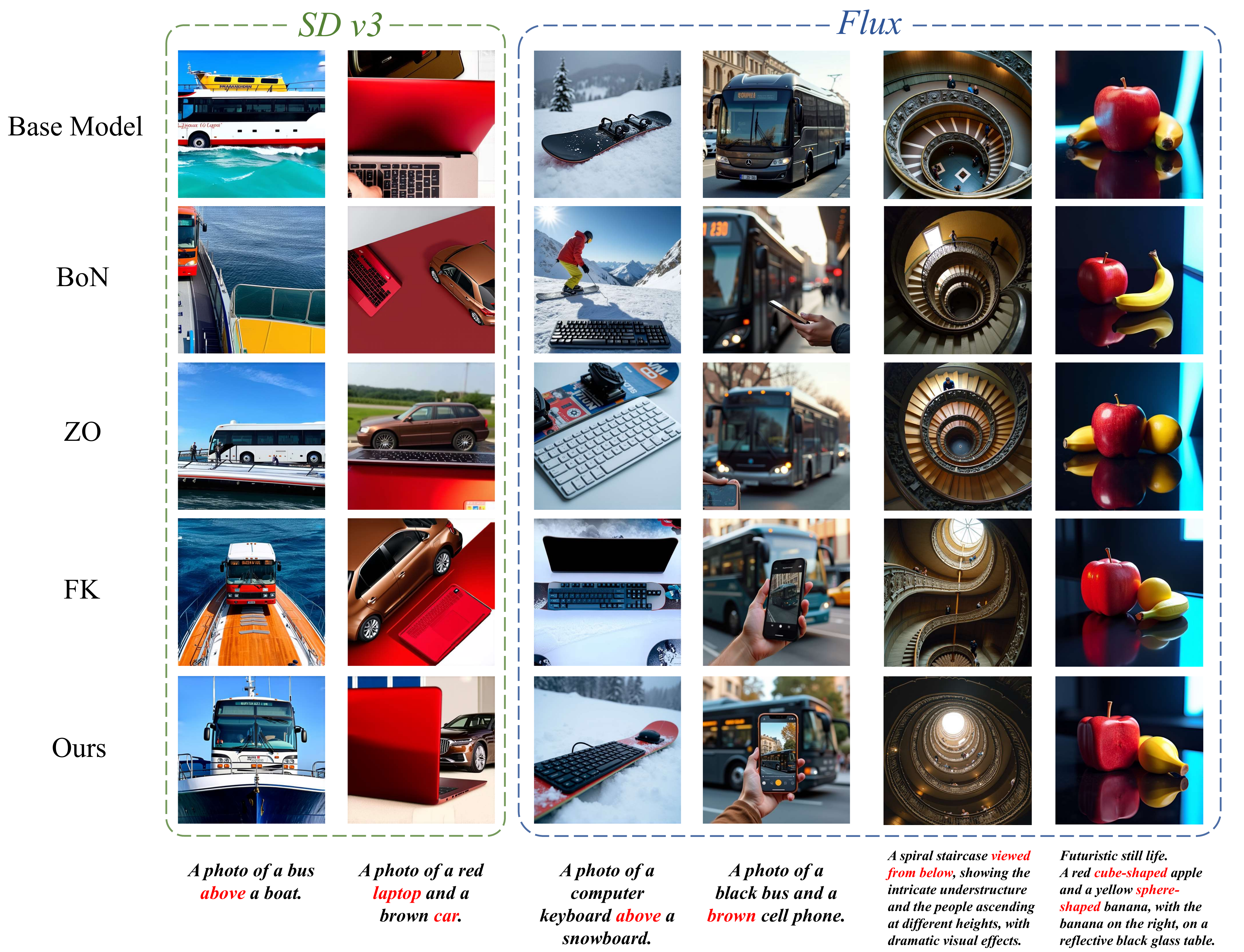} 
\caption{Visualizations of text-to-image generation. Building on FLUX and SD v3 as backbones, we apply our method and multiple baselines to generate images conditioned on diverse prompts from GenEval and GPT-4. All TTS methods employ the same ImageReward verifier during optimization.}
\label{fig2:example}
\vspace{-0.2in}
\end{figure*}

\subsection{Main Results}
\textbf{Quantitative evaluation.} We first report the quantitative results on the GenEval dataset in Table~\ref{tab1}. Our method outperforms all competitors across all held-in and held-out metrics. Compared to the base models (without TTS), our approach delivers substantial gains across all metrics, notably boosting held-out GenEval by 20.7\% (SD) / 15.6\% (FLUX) and held-in ImageReward by 66.0\% (SD) / 60.5\% (FLUX), alongside consistent improvements in NIQE and CLIP. Furthermore, compared to the strongest baseline (FK), our RTS yields consistent improvements in both GenEval and ImageReward across SD (4.4\% / 8.7\%), FLUX (4.6\% / 7.5\%), and Qwen-Image (6.5\% / 5.1\%), while remarkably reducing the latter's inference time by 31.9 seconds. Similarly, our RTS outperforms all competitors across all tasks in the T2I-CompBench benchmark, as shown in Table~\ref{tab:t2i_compbench}.

To demonstrate that our method is not overly reliant on ImageReward, we conduct an additional comparison against Noise Trajectory Search (NTS). Following the official NTS implementation, we employ the Stable Diffusion v1.5 backbone and utilize a CLIP verifier for search guidance rather than ImageReward. As shown in Table~\ref{CLIP}, our method achieves comprehensive improvements across all held-in and held-out metrics, while reducing inference time by over $50\%$ (107s vs. 52s) under the same NFE budget. This confirms that our method's superiority is agnostic to the chosen verifier.

\textbf{Qualitative evaluation.} We conduct qualitative experiments to evaluate our method against various competitors on prompt complexities. As shown in Figure~\ref{fig2:example}, compared to baselines, our approach consistently achieves superior semantic alignment and visual fidelity. For instance, in the first prompt, our RTS accurately captures the spatial relationship (bus above boat) with a more harmonious composition. While both our method and the ZO baseline correctly identify the ``brown cellphone" in the fourth prompt, ours produces significantly more refined textures.  Furthermore, our RTS excels at following fine-grained details in complex prompts, such as the specific perspective in the fifth prompt and the counter-intuitive fruit shapes in the sixth prompt. These results highlight the effectiveness and robustness of our RTS in enhancing the generation performance of diffusion models. We provide more visual comparisons in Appendix Figure~\ref{fig7}.

\begin{table*}[t]
\centering
\caption{Quantitative results on the T2I-CompBench benchmark across six tasks. All TTS methods employ the same ImageReward verifier during optimization.}
\label{tab:t2i_compbench}
\resizebox{0.85\linewidth}{!}{
\begin{tabular}{lcccccccc}
\toprule
{Base Model} 
& {TTS Baseline} 
& Color 
& Shape 
& Texture 
& Spatial 
& Non-spat
& Complex 
& {NFEs}  \\
\midrule

\multirow{5}{*}{Flux-1.dev}
& TTS Free   & 0.7692 & 0.5187 & 0.6287 & 0.2429 & 0.2694 & 0.3600 & 50   \\
& BoN~\citep{ma2025inference}  & 0.8203 & 0.6274 & 0.7364 & 0.3151 & 0.2856 & 0.3810 & 1000 \\
& ZO~\citep{ma2025inference}  & 0.8311 & 0.6296 & 0.7441 & 0.3189 & 0.2934 & 0.3897 & 1000 \\
& FK~\citep{singhal2025general}  & 0.8474 & 0.6343 & 0.7486 & 0.3224 & 0.3017 & 0.3989 & 1000 \\
& Ours & \textbf{0.8723} & \textbf{0.6474} & \textbf{0.7630} 
       & \textbf{0.3431} & \textbf{0.3245} & \textbf{0.4101} & 1000 \\

\bottomrule
\end{tabular}
}
\vspace{-0.1in}
\end{table*}

\begin{table*}[t]
\centering
\caption{Quantitative comparison between NTS and our method on the GenEval benchmark. This confirms that our performance gains do not overly rely on ImageReward as the verifier.}
\resizebox{0.88\linewidth}{!}{
\begin{tabular}{lllcccccc}
\toprule 
\multirow{2}{*}{Method} & \multirow{2}{*}{Backbone} & \multirow{2}{*}{Verifier} & Held-in Metric & \multicolumn{3}{c}{Held-out Metrics} & \multirow{2}{*}{NFEs} & \multirow{2}{*}{Runtime (s)} \\
\cmidrule(r){4-4} \cmidrule(l){5-7} 
& & & CLIP $\uparrow$ & ImageReward $\uparrow$ & GenEval $\uparrow$ & NIQE $\downarrow$ & & \\ 
\midrule

NTS~\citep{ramesh2025test} & \multirow{2}{*}{SDv1.5} & \multirow{2}{*}{CLIP} 
& 0.2573 & 0.2658 & 0.3375 & 4.2361 & 1000 & 107 \\

Ours & &  
& \textbf{0.2811} & \textbf{0.4556} & \textbf{0.4327} & \textbf{3.9804} & 1000 & 52 \\

\bottomrule
\end{tabular}
}
\label{CLIP}
\vspace{-0.1in}
\end{table*}

\begin{table*}[t]
\centering
\caption{Ablation studies of the initial
noise and intermediate noise optimization in the GenEval benchmark dataset under a fixed NFE budget. The first row represents the base model without search. Additional ablations under fixed search parameters (varying NFEs) are put on Appendix Table~\ref{varing}.}
\label{tab2-1}
\resizebox{0.85\linewidth}{!}{
\begin{tabular}{lccccccc}
\hline
Base Model & Initial Noise & Intermediate Noise & ImgReward  $\uparrow$ &  GenEval $\uparrow$ & NIQE $\downarrow$ & CLIP $\uparrow$ & NFEs \\ \hline
\multirow{4}{*}{Flux-1.dev} &  &  & 0.9159 & 0.6516 & 4.2171 & 0.2727 & 50 \\
  & $\checkmark$ &  &  1.4201 & 0.7412 & 3.9722 & 0.2861 & 1000 \\
 &  & $\checkmark$ &  1.4183 & 0.7319 & 4.0054 & 0.2937 & 1000 \\
 
 & $\checkmark$ & $\checkmark$ &  \textbf{1.4703} & \textbf{0.7535}  & \textbf{3.8506} & \textbf{0.2964} & 1000 \\ \hline
\end{tabular}
}
\label{tab2-2}
\end{table*}

\subsection{Analysis and Discussions}

\textbf{Effect of the initial and intermediate noise optimization.} We conduct comprehensive ablation studies to assess the effect of the distinct components within RTS. Table~\ref{tab2-1} highlights that: 1) both the initial and intermediate optimization stages secure substantial performance gains; and 2) the integrated framework demonstrates strong synergy, fully exploiting the potential of the noise space.

\textbf{Discussion of the initial noise optimization.} Our initial noise optimization consists of two search strategies: coarse-grained and fine-grained. We further conduct ablation experiments to explore their effects. As reported in Table~\ref{ini_qua}, while coarse search alone effectively boosts ImageReward through broad exploration, other metrics (\emph{i.e.}, GenEval, NIQE, and CLIP) remain relatively stable without targeted refinement. Adding the fine-grained search perfectly bridges this gap. This entire coarse-to-fine mechanism unlocks the best overall performance across all metrics, proving that broad exploration and detailed refinement are both indispensable.

\textbf{Discussion on the sparse test-time scaling.} To further evaluate our sparse scaling, we compare it against a variant that performs search across all denoising steps (cf. Table~\ref{secnd_aba}). Notably, full scaling degrades performance (e.g., lowering ImgReward). We attribute this to the phase-sensitive nature of diffusion models: early noisy latents establish the overall object structures~\citep{meng2021sdedit,si2024freeu,fu2024generate}. Forcing scaling interventions during these sensitive phases disrupts the generation trajectory. Instead, by leveraging PCA to analyze representation variation rates, we identify and exclusively target a sparse set of mid-stage inflection points (cf. Figure~\ref{pca_points}). This targeted strategy effectively avoids trajectory collapse and ultimately boosts the generation performance.

\textbf{Ablation on key-timestep selection.} We validate our PCA-based step selection against two alternatives: 1) \textit{Uniform step selection}, which selects equidistant steps across the entire trajectory; and 2) \textit{Reward-sensitive adaptive step selection}, which targets steps with the highest variation rates in ImageReward scores. As shown in Table~\ref{pca_aaa}, the reward-based selection yields suboptimal results, likely because reward scores are highly sensitive to changes in overall object structures, causing the selected steps to cluster primarily in the early phase. Scaling interventions during this sensitive initial denoising phase disrupts the generation trajectory. In contrast, our PCA-curvature outperforms uniform by precisely targeting key mid-stage inflection points (cf. Figure~\ref{pca_points}). More ablations on random and fixed-stage (early, mid, late) selection strategies are provided in the Appendix~\ref{sparse-time}.

\textbf{Wall-clock/quality Pareto efficiency.} Regarding wall-clock efficiency, our RTS forms a strictly superior Pareto frontier against existing TTS baselines (e.g., BoN, ZO, and FK) under equal runtime budgets (cf. Appendix Table~\ref{tab:runtime_comparison}). Furthermore, compared to standard single-pass (TTS-free) inference, although RTS intrinsically incurs additional sampling time, it offers a crucial paradigm shift by trading inference time for VRAM efficiency, thus empowering VRAM-constrained users with high-quality generation capabilities. Further discussions are provided in the Appendix~\ref{Wall-clock}.

\begin{table}[t]
\begin{minipage}{0.5\linewidth}
\centering
    \caption{Effect of the two search strategies within the initial noise optimization on the FLUX model at fixed 1000 NFEs budget. The first row denotes the RTS variant without initial noise optimization.}
    \vspace{0.1in}

\resizebox{1.0\linewidth}{!}{
\begin{tabular}{ccccccc}
\hline
 Coarse & Fine &  ImgReward $\uparrow$ & GenEval  $\uparrow$ & NIQE $\downarrow$ & CLIP $\uparrow$  \\ \hline
  & & 1.4183  & 0.7319 & 4.0054 & 0.2937  \\
$\checkmark$ &   &  1.4403  & 0.7315 & 4.0248 & 0.2893  \\
  $\checkmark$ &  $\checkmark$ & \textbf{1.4703} &  \textbf{0.7535}& \textbf{3.8506} & \textbf{0.2964}\\ \hline
\end{tabular}
}
\label{ini_qua}

\end{minipage}
\hfill
\begin{minipage}{0.47\linewidth}
\centering

\caption{Effect of the sparse scaling strategy on FLUX at  1000 NFEs. ``Base'' denotes the RTS variant utilizing initial noise optimization but omitting the intermediate noise optimization.}
\vspace{0.1in}
\resizebox{0.97\linewidth}{!}{
\begin{tabular}{ccccc}
\hline
 Scaling  & ImgReward $\uparrow$ &  GenEval$\uparrow$ & NIQE $\downarrow$ & CLIP $\uparrow$  \\ \hline
                       Base         &       1.4201      &  0.7412             &   3.9722       &    0.2861    \\
  Full &  1.4072 & 0.7372 & 4.0133 & 0.2876  \\

   Sparse & \textbf{1.4703}   & \textbf{0.7535}   & \textbf{3.8506}  & \textbf{0.2964}  \\ \hline
\end{tabular}
}
\label{secnd_aba}

\end{minipage}
\end{table}

\begin{table}[H]
\begin{minipage}{0.52\linewidth}
\caption{Key-timestep selection ablations on the FLUX model at a fixed 1000 NFEs budget.}
\vspace{0.1in}
\label{pca_aaa}
\centering
\resizebox{1.0\linewidth}{!}{
\begin{tabular}{ccccc}
\hline
Key-step selection & ImgReward$\uparrow$ & GenEval$\uparrow$ & NIQE $\downarrow$   & CLIP  $\uparrow$  \\ \hline
Reward-sensitive    & 1.4203  &  0.7366  & 4.0703 & 0.2869  \\
Uniform     & 1.4404   & 0.7419   & 4.0435 & 0.2846  \\
Ours (PCA-curvature)   &  \textbf{1.4703} & \textbf{0.7535}   & \textbf{3.8506} & \textbf{0.2964} \\ \hline
\end{tabular}}

\end{minipage}
\hfill
\begin{minipage}{0.44\linewidth}
\centering
\caption{Ablation study of different verifiers on the FLUX model at 1000 NFEs.}
\vspace{0.1in}
\label{tab:my-table}
\resizebox{0.97\linewidth}{!}{\begin{tabular}{ccccc}
\hline
Verifier &  ImgReward$\uparrow$ & GenEval $\uparrow$ & NIQE $\downarrow$  & CLIP $\uparrow$  \\ \hline
\begin{tabular}[c]{@{}c@{}}LLMGrader\end{tabular}  &  1.2949 & 0.7363 & 4.1715 & 0.2647  \\
\begin{tabular}[c]{@{}c@{}}CLIP\end{tabular}      & 1.2514 & 0.7336 & 4.1593 & \textbf{0.3182}  \\
\begin{tabular}[c]{@{}c@{}}ImgReward\end{tabular} & \textbf{1.4703} & \textbf{0.7535} & \textbf{3.8506} & 0.2964  \\ \hline
\end{tabular}}
\label{verifier}

\end{minipage}
\end{table}

\section{Conclusion}
\label{sec:conclu}

In this work, we introduce test-time scaling as an effective paradigm for enhancing diffusion models, demonstrating that computational allocation through comprehensive exploration of the noise space significantly improves sample quality across architectures and metrics. Our proposed sparse test-time scaling framework unifies initial noise optimization and intermediate noise optimization, achieving state-of-the-art performance on the popular GenEval and T2I-CompBench benchmarks while maintaining favorable computational cost. Ultimately, our framework provides both a practical solution and valuable insights for test-time scaling in diffusion models.

\clearpage

\bibliography{iclr2027_conference}
\bibliographystyle{iclr2027_conference}

\clearpage
\appendix

\begin{table}
	\setlength{\tabcolsep}{0.2cm}
	\begin{tabular}{p{0.97\columnwidth}}
		\centering
		\textbf{\large{Guided Denoising Trajectory Optimization for Test-Time Diffusion Scaling \\ \vspace{0.1in}Supplementary Materials}}
	\end{tabular}
\end{table}


\section{More Related Work}
\label{sec:more_related}
\textbf{Zeroth-Order Search}
aims to perform gradient-based search without accessing explicit gradients, a capability crucial for scenarios including optimizing black-box systems, tuning models with non-differentiable components, and working in privacy-sensitive or hardware-constrained environments. 

Its versatility has led to broad applications in machine learning~\citep{liu2020primer} and natural language processing~\citep{malladi2023fine}, with notable methods including BBT~\citep{sun2022black}, BBTv2~\citep{sun2022bbtv2}, and RLPrompt~\citep{deng2022rlprompt}. However, these approaches often suffer from challenges related to high variance and instability~\citep{liu2020primer}. To address these limitations, recent work has focused on improving gradient estimation accuracy via techniques such as two-sided approximation~\citep{oh2023blackvip} and reducing variance in Zeroth-Order (ZO) fine-tuning of large language models through sparse parameter perturbation~\citep{liu2024sparse}. Other lines of work optimize ZO itself by incorporating historical data~\citep{cheng2021convergence} and reusing intermediate features~\citep{chen2023deepzero}. These efforts collectively demonstrate that ZO methods can handle complex tasks with notable adaptability and efficiency.

ZO-based methods offer a promising pathway for test-time scaling due to their capability for gradient-free search. 
However, prior methods have primarily treated ZO search as a simple filtering strategy, which, though straightforward, offers limited gains. In contrast, we propose a coarse-to-fine zeroth-order search scheme. Through an alternating-round design, our approach balances exploration in the noise state space with the effective utilization of historical information.

\textbf{Diffusion models.} Denoising Diffusion Probabilistic Model~\citep{ho2020denoising} establish a powerful paradigm for image generation~\citep{cao2019multi,nichol2021glide,ruiz2023dreambooth,liu2024implicit}. To reduce the computational costs, Latent Diffusion Model~\citep{rombach2022high} shifts the diffusion process to a compressed latent space. Parallel to these architectural advancements, control mechanisms have evolved from early classifier-based guidance~\citep{dhariwal2021diffusion} to the more integrated classifier-free guidance~\citep{ho2022classifier}. Building on these foundations, recent text-to-image models such as DALL-E3~\citep{betker2023improving}, Stable Diffusion~\citep{esser2024scaling}, and FLUX~\citep{batifol2025flux} leverage prompts to guide the diffusion process, producing highly realistic and diverse generation.

\section{More implementation details}
\label{sec:imp}

\subsection{Definition of local curvature}
\label{sec:loca}
Our curvature calculation follow the established framework of Discrete Differential Geometry~\citep{crane2018discrete}. For a discrete trajectory sequence defined as $P = \{p_1, p_2, \dots, p_N\} \in \mathbb{R}^{N \times d}$, we first approximate the first derivative $r'_i$ (velocity/tangent) and second derivative $r''_i$ (acceleration/curvature trend) at step $i$ using finite differences:
\begin{equation}
\begin{gathered}
r'_i = \frac{p_{i+1} - p_{i-1}}{2}, \\
r''_i = p_{i+1} - 2p_i + p_{i-1}.
\end{gathered}
\end{equation}

The local curvature $\kappa_i$ is then calculated as:
\begin{equation}
\begin{gathered}
\kappa_i = \frac{\sqrt{\Vert{}r'_i\Vert{}^2 \Vert{}r''_i\Vert{}^2 - (r'_i \cdot r''_i)^2}}{\Vert{}r'_i\Vert{}^3}.
\end{gathered}
\end{equation}
Because this formulation relies exclusively on vector norms and dot products, it is mathematically valid for any arbitrary dimension $d$.

\subsection{Pseudo code of the proposed reward-guided noise optimization}
\label{pseudo}
\begin{algorithm}[H]
\caption{Reward-Guided Noise Optimization}
\label{alg:adaptive_zo}
\begin{algorithmic}[1]
\REQUIRE Neighbor count $N$, search round $T$, reward function $R$, denoising step $L$, weight $\alpha$
\FOR{$t = 1$ to $T$}
        \IF{$t \pmod 2 \equiv 1$}
        \STATE \COMMENT{Coarse-grained search}
        \IF{$t > 1$ \AND $\max(\mathcal{R}_{t-1}) > R(\mathbf{s}_{t-1})$}
            \STATE $\mathbf{s}_{t} \gets \arg\max_{\mathbf{m} \in \mathcal{M}_{t-1}} R(\mathbf{m})$
        \ELSE
            \STATE $\mathbf{s}_{t} \sim \mathcal{N}(0, I)$
        \ENDIF
        \STATE $\mathcal{M}_{t}, \mathcal{W}_t \small\gets \text{Random Spherical Sampling}(\mathbf{s}_{t}, N)$ %
        \STATE $\mathcal{R}_t \gets \text{Denoising and scoring}(\mathcal{M}_{t}, L, R)$ 
        \STATE $\mathbf{g_t} \gets \text{Gradient Estimation} (R(\mathbf{s}_{t}), \mathcal{R}_t, \mathcal{W}_t)$
        
    \ELSE 
        \STATE \COMMENT{Fine-grained search}
        \STATE $\mathbf{g}^\perp \gets \text{OrthogonalProjection}(\mathbf{g_{t-1}}, \mathbf{s}_{t})$ 
        \STATE $\mathcal{W}_t \gets (1-\alpha)\mathcal{W}_{t-1} + \alpha\mathbf{g}^\perp$
        \STATE $\mathcal{M}_{t} \gets \text{Guided Spherical Sampling}(\mathbf{s}_{t}, N, \mathcal{W}_t)$ 
        \STATE $\mathcal{R}_t \gets \text{Denoising and scoring}(\mathcal{M}_{t}, L, R)$ 
    \ENDIF
\ENDFOR
\STATE $\mathbf{m}_T^* \gets \arg\max_{\mathbf{m} \in \mathcal{M}_{T}} R(\mathbf{m})$ 
\ENSURE Optimized noise $\mathbf{m}_T^*$
\end{algorithmic}
\end{algorithm}

\subsection{Interpolation choice of Equation~(\ref{guided_qph})} 
\label{choice}
Regarding the choice of linear interpolation $(1-\alpha, \alpha)$, the core reason we utilize standard linear interpolation here, rather than a norm-preserving one, is that our update rule intentionally separates direction estimation from sample generation into two distinct geometric spaces : 

\textbf{Direction estimation in the tangent space.} The calculation $\hat{\mathbf{w}}_i = (1 - \alpha) \cdot \hat{\mathbf{w}}_i' + \alpha \cdot \frac{\mathbf{g}^\perp}{\Vert{}\mathbf{g}^\perp\Vert{}}$ occurs strictly within the local tangent space orthogonal to the base latent direction $\mathbf{u}$. Because this tangent plane is locally Euclidean, a standard linear combination is the mathematically natural to aggregate vectors (i.e., blending the historical momentum with the optimal ascent direction).

\textbf{Sample generation on the hypersphere.} Conversely, when generating the actual neighbor latent $\mathbf{m}^i$, the point must be mapped back onto the hypersphere to preserve the original norm of the high-dimensional latents, as generative models are highly sensitive to the magnitude of latent vectors. A naive linear addition to the base latent would alter this norm, undesirably conflating directional optimization with latent magnitude distortion. Therefore, as shown in the second line of Equation~(\ref{guided_qph}), we employ a norm-preserving spherical formulation:
$$\mathbf{m}^i = r \cdot \left( \tau \cdot \mathbf{u} + \sqrt{1 - \tau^2} \cdot \frac{\hat{\mathbf{w}}_i}{\Vert{}\hat{\mathbf{w}}_i\Vert{}} \right)$$

This step explicitly extracts the direction from the tangent plane, normalizes it by its magnitude ($\Vert{}\hat{\mathbf{w}}_i\Vert{}$), and projects it onto the sphere. This guarantees that the newly generated latent maintains a strict angular distance (governed by $\tau$) from the base latent $\mathbf{u}$.

In summary, the linear interpolation $(1-\alpha, \alpha)$  aggregates directions on the Euclidean tangent plane, while the subsequent equation strictly handles the norm-preserving mapping onto the sphere. Introducing a norm-preserving interpolation during the initial blending step is mathematically redundant, as the intermediate magnitude of the interpolated $\hat{\mathbf{w}}_i$ is intentionally discarded prior to the final spherical projection.

\subsection{Formulation details of Equation~(\ref{est_grad})} 
\label{Derivation}
Equation~(\ref{est_grad}) formulates a zero-order surrogate gradient $\mathbf{g}$ that synthesizes robust directional guidance without requiring backpropagation. Specifically, it aggregates differential signals from neighbor candidates, which can be understood through the following three aspects:

\textbf{Relative difference Estimation}: The term $R(\mathbf{m}_i) - R(\mathbf{z})$ computes the relative reward difference between a perturbed neighbor sample and the base latent. By utilizing this relative difference rather than absolute scores, the formulation intrinsically evaluates whether a specific tangential perturbation direction $\hat{\mathbf{w}}_i$ leads to a localized improvement in the reward landscape.

\textbf{Normalization}: The relative advantage is divided by the total local reward mass, $\sum_{i=1}^N R(\mathbf{m}_i) + R(\mathbf{z})$. This acts as a normalizer to stabilize the gradient estimation. It ensures that the guidance remains robust, preventing extreme absolute reward values from causing numerical instability.

\textbf{Orthogonal directional aggregation}: Equation~(\ref{est_grad}) finally aggregates these normalized, reward-weighted signals along the unit orthogonal vectors $\hat{\mathbf{w}}_i$. This linear combination accurately constructs the steepest ascent direction $\mathbf{g}$ entirely within the local Euclidean tangent plane.

By leveraging these evaluation priors on neighboring latents, this formulation effectively identifies and actively explores regions with high reward potential.

\subsection{Intermediate-noise optimization procedure}
\label{intermediate}
As illustrated in the Figure~\ref{fig1:method}, the procedure can be broken down into the following distinct steps:

\textbf{Path generation and latent scoring.} The optimized initial noise is first input into the SDE-solver to generate four parallel denoising paths. Along these paths, at each key-step, after modifying the four intermediate latents, we pass them through the SDE-solver to execute a single-step denoising. This step predicts the clean images (e.g., $x_0$), which are then directly evaluated by the reward model to obtain their respective performance scores.

\textbf{Latent resampling and resuming.} We utilize these predicted scores as weight factors for the four intermediate latents and perform a resampling strategy to obtain four new intermediate latents. This resampling strategy ensures that promising intermediate latents are highly likely to be sampled multiple times (duplicated), whereas poor latents have a probability of being terminated. These four newly sampled latents are then fed back into the SDE-solver to seamlessly execute the subsequent denoising steps.

\textbf{Trajectory Resumption.} We input the modified latents back into the SDE-solver to compute fresh predictions for the subsequent steps. Consequently, the modified latent acts purely as a mathematically valid new starting state for the next denoising step.

\subsection{More hyperparameter and inference details}
\label{hyperparameter}
\textbf{Hyperparameter details.} Our hyperparameter configuration ($k, N, T, \alpha, \tau$) were initially determined using the FLUX architecture on the GenEval benchmark. Notably, for the other architectures evaluated on GenEval (i.e., SD3 and QwenImage), we directly applied these FLUX-derived parameters without any additional tuning. To ensure these choices were robust, they were subsequently evaluated on the T2I-CompBench benchmark, which served as a strictly held-out prompt set.

Specifically, our validation protocol is: 1) Search Budget Parameters ($k, N, T$): We conducted a comprehensive parameter sensitivity analysis on the GenEval benchmark using FLUX architecture. The empirical results in Tables~\ref{tab:key_steps_ablation} and~\ref{tab:ablation_horizontal} demonstrate that our method is highly robust and not sensitive to these search budget parameters, with performance remaining consistently stable across a wide range of values. Consequently, to strike a balance between performance and computational overhead, we selected relatively small parameter values. 2) Balancing Factors ($\alpha, \tau$) : These components were determined via a standard grid search on the GenEval benchmark using FLUX architecture.

\textbf{Inference details.} During inference, we utilize the SDE-based scheduler paired with the FlowMatch-Heun sampler~\citep{esser2024scaling} at a 512×512 resolution. We follow the default hyperparameters provided in official codes: 28 denoising steps and a guidance scale of 7.0 for SD3, 50 denoising steps with a guidance scale of 3.5 for FLUX , and 50 denoising steps with a guidance scale of 4.0.

\begin{table}[t]
\centering
\caption{Ablation study on the number of key steps $k$ at 1000 NFEs.}
\label{tab:key_steps_ablation}
\setlength{\tabcolsep}{9pt}
\renewcommand{\arraystretch}{1.12}
\resizebox{0.7\linewidth}{!}{
\begin{tabular}{lccccc}
\toprule
Backbone & $k$ & GenEval $\uparrow$ & IR $\uparrow$ & NIQE $\downarrow$ & CLIP $\uparrow$ \\
\midrule

\multirow{4}{*}{Flux-1.dev}
& 3  & 0.7418 & 1.4374 & 4.0047 & 0.2877 \\
& 6  & 0.7535 & 1.4703 & 3.8506 & 0.2964 \\
& 9  & 0.7541 & 1.4883 & 3.8717 & 0.2937 \\
& 12 & 0.7547 & 1.4907 & 3.8178 & 0.3011 \\

\bottomrule
\end{tabular}
}
\end{table}

\begin{table*}[t] 
    \centering
    \footnotesize
    \caption{\textbf{Computational investment analysis.} Comparisons of different budget allocations.}
    \label{tab:ablation_horizontal}
    
    \begin{subtable}{0.32\linewidth} 
        \centering
        \caption{Effect of Round Params ($N_1$)}
        \resizebox{\linewidth}{!}{
            \begin{tabular}{cccc} 
                \toprule
                $N_1$ & GenEval$\uparrow$ & ImgReward$\uparrow$ & NFEs \\ 
                \midrule
                2  & 0.7404  & 1.4149    & 600  \\ 
                4  & 0.7535  & 1.4703    & 1000 \\ 
                8  & 0.7592  & 1.5197    & 1600 \\ 
                \bottomrule
            \end{tabular}
        }
    \end{subtable}
    \hfill 
    \begin{subtable}{0.32\linewidth}
        \centering
        \caption{Effect of Nghbr Noise ($K_1$)}
        \resizebox{\linewidth}{!}{
            \begin{tabular}{cccc}
                \toprule
                $K_1$ & GenEval$\uparrow$ & ImgReward$\uparrow$ & NFEs \\ 
                \midrule
                2  & 0.7373  & 1.3984    & 600  \\ 
                4  & 0.7535  & 1.4703    & 1000 \\ 
                8  & 0.7563  & 1.502     & 1600 \\ 
                \bottomrule
            \end{tabular}
        }
    \end{subtable}
    \hfill 
    \begin{subtable}{0.32\linewidth}
        \centering
        \caption{Effect of Opt Paths ($N_2$)}
        \resizebox{\linewidth}{!}{
            \begin{tabular}{cccc}
                \toprule
                $N_2$ & GenEval$\uparrow$ & ImgReward$\uparrow$ & NFEs \\ 
                \midrule
                2  & 0.7433  & 1.4499    & 900  \\
                4  & 0.7535  & 1.4703    & 1000 \\
                8  & 0.7587  & 1.4841    & 1200 \\ 
                \bottomrule
            \end{tabular}
        }
    \end{subtable}
\end{table*}

\section{User preference study}
\label{sec:user}
We conduct two independent blind human preference studies to evaluate the FLUX and Qwen-Image architectures separately. In each study, participants were shown a text prompt along with candidate images generated by different methods, and were asked to select the image that best aligned with the prompt. For the FLUX-based methods, we collected 400 valid responses from 20 volunteers. We conducted a separate evaluation for the Qwen-Image-based methods, collecting an additional 400 valid responses from 20 volunteers.

As shown in Table~\ref{tab:user_study}, our method achieves consistent and significant gains not just on a single metric, but across all held-out evaluators. We emphasize that these improvements remain exceptionally clear under evaluators that were never used to guide the search. The blind human preference studies confirm genuine perceptual improvements: our approach outperforms FK in user preference by a $14.5\%$ margin on FLUX-1-dev and $19.0\%$ on Qwen-Image, comprehensively validating its effectiveness across both architectures.

\begin{table}[H]
\centering
\caption{User preference evaluation across different TTS methods.}
\label{tab:user_study}
\setlength{\tabcolsep}{12pt}
\renewcommand{\arraystretch}{1.12}
\resizebox{0.7\linewidth}{!}{
\begin{tabular}{lcccc}
\toprule
Backbone & BoN & ZO & FK & \textbf{Ours} \\
\midrule
Flux-1.dev & 20.75\% & 14.25\% & 25.25\% & \textbf{39.75\%} \\
Qwen-Image & $-$ & $-$ & 40.50\% & \textbf{59.50\%} \\
\bottomrule
\end{tabular}
}
\end{table}

\section{Extended Analysis and Discussions}
\label{sec:more_analysis}

\subsection{Scaling Behavior} We analyze the effect of computational scaling via three control parameters: alternating rounds (N1), neighborhood noise size (K1), and the number of optimized trajectory (N2). As shown in Tables~\ref{tab:ablation_horizontal} (a)–(b), scaling up these parameters results in a steady improvement in GenEval and ImageReward metrics. The growth pattern aligns with established scaling laws, showing clear performance benefits from increased compute. Crucially, our framework demonstrates exceptional versatility: even at conservative parameter settings with lower NFEs, it yields results superior to baselines. This indicates that our approach allows for a flexible trade-off between computational cost and generation quality, making it effective across a wide range of resource constraints.

\begin{table*}[t]
\centering
\caption{Ablation studies of the initial
noise and intermediate noise optimization in the GenEval benchmark dataset under fixed search parameters.}
\label{varing}

\resizebox{0.98\linewidth}{!}{
\begin{tabular}{lccccccc}
\hline
Base Model & Initial Noise & Intermediate Noise & ImgReward  $\uparrow$ &  GenEval $\uparrow$ & NIQE $\downarrow$ & CLIP $\uparrow$ & NFEs \\ \hline
\multirow{4}{*}{Flux-1.dev} &  &  & 0.9159 & 0.6516 & 4.2171 & 0.2727 & 50 \\
  & $\checkmark$ &  &  1.4085 & 0.7389 & 4.0435 & 0.2849 & 800 \\
 &  & $\checkmark$ &  1.4102 & 0.7261 & 4.0157 & 0.2886 & 200 \\
 
 & $\checkmark$ & $\checkmark$ &  \textbf{1.4703} & \textbf{0.7535}  & \textbf{3.8506} & \textbf{0.2964} & 1000 \\ \hline
\end{tabular}
}
\end{table*}

\subsection{Reward model ablations} The choice of reward model is critical, as it determines the direction of trajectory optimization during test-time scaling. To evaluate this, we conduct ablation experiments using several distinct verifiers: 2) LLMGrader, which leverages large language models for automated assessment; 2 metric-based verifiers, including CLIP and ImageReward.
As shown by the quantitative results in Table~\ref{verifier}, ImageReward verifier consistently achieves the best performance across most metrics, including GenEval, ImageReward, and NIQE, while maintaining competitive scores on CLIP. Consequently, we select ImageReward as our primary reward model for the final implementation.

\begin{table}[t]
\centering
\caption{Ablation study on different PCA dimensions on the GenEval dataset.}
\label{tab:pca_dimension}
\setlength{\tabcolsep}{10pt}
\renewcommand{\arraystretch}{1.12}
\resizebox{0.7\linewidth}{!}{
\begin{tabular}{lcccc}
\toprule
Dimension & ImageReward $\uparrow$ & GenEval $\uparrow$ & NIQE $\downarrow$ & CLIP $\uparrow$ \\
\midrule
PCA-3 & \textbf{1.4703} & \textbf{0.7535} & \textbf{3.8506} & \textbf{0.2964} \\
PCA-4 & 1.4583 & 0.7462 & 3.8958 & 0.2917 \\
PCA-5 & 1.4436 & 0.7417 & 3.9149 & 0.2935 \\
\bottomrule
\end{tabular}
}
\end{table}

\subsection{Choice of projection dimension}
\label{pca-ablation}
The projection dimension of 3 was selected because we found that the first three principal components already capture the most significant variance of the entire denoising trajectory. Specifically, we calculated the Cumulative Explained Variance Ratio on the GenEval benchmark and observed that the first three components retain $99.93\%$ of the trajectory information. This demonstrates that three dimensions are entirely sufficient for accurately representing the trajectory.

To further validate this, we provide a sensitivity analysis across different PCA dimensions, as shown in Table~\ref{tab:pca_dimension}. The results indicate that the overall performance difference among varying dimensions is minimal. However, PCA-3 achieves slightly better results across several metrics, particularly ImageReward and GenEval. One potential reason for this marginal improvement is that restricting the projection to 3 dimensions effectively filters out redundant information from the trajectory.

\begin{table}[H]
\centering
\caption{Comparison of different keypoint calculation strategies.}
\label{tab:keypoint_strategy}
\setlength{\tabcolsep}{8pt}
\renewcommand{\arraystretch}{1.12}
\resizebox{\linewidth}{!}{
\begin{tabular}{lcccccc}
\toprule
Method & ImageReward $\uparrow$ & GenEval $\uparrow$ & NIQE $\downarrow$ & CLIP $\uparrow$ & NFEs & Runtime \\
\midrule
Per-Round Recalculation & 1.4649 & \textbf{0.7579} & 3.8964 & 0.2950 & 1110 & 244.1 \\
Ours (Shared initial keypoints) & \textbf{1.4703} & 0.7535 & \textbf{3.8506} & \textbf{0.2964} & \textbf{1000} & 217.7 \\
\bottomrule
\end{tabular}
}
\end{table}

\subsection{Disscusions about the roubutness of high-curvature keypoints}
\label{roubutness-key}
Whether the altered trajectory is significant enough to invalidate the previously selected points, we found these keypoints to be highly resilient. To verify this, we conduct ablation experiments comparing two versions: sharing the initial set of keypoints across all optimization rounds versus recalculating them in every single round. As shown in Table~\ref{tab:keypoint_strategy}, we observe that both of them yield comparable performance as shown in the table below, which indicates that the high-curvature keypoints are robust to the stochastic variations introduced by the altered trajectories. Because sharing the initial keypoints across all rounds is significantly more time-efficient, we ultimately chose this shared version for our implementation.

\begin{table}[t]
\centering
\caption{Comparison of different key-step selection strategies.}
\label{tab:key_step_selection}
\setlength{\tabcolsep}{10pt}
\renewcommand{\arraystretch}{1.12}
\resizebox{0.8\linewidth}{!}{
\begin{tabular}{lcccc}
\toprule
Key-step selection & ImageReward $\uparrow$ & GenEval $\uparrow$ & NIQE $\downarrow$ & CLIP $\uparrow$ \\
\midrule
Random & 1.4217 & 0.7368 & 4.0660 & 0.2863 \\
Fixed: Early & 1.4078 & 0.7296 & 4.1412 & 0.2763 \\
Fixed: Middle & 1.4516 & 0.7479 & 4.0198 & 0.2924 \\
Fixed: Late & 1.4214 & 0.7411 & 3.9784 & 0.2864 \\
Ours & \textbf{1.4703} & \textbf{0.7535} & \textbf{3.8506} & \textbf{0.2964} \\
\bottomrule
\end{tabular}
}
\end{table}

\subsection{More ablations on sparse timestep selection}
\label{sparse-time}
To provide a more comprehensive analysis of our PCA-based sparse timestep selection, we evaluate it against several heuristic baselines under a strictly controlled computational budget (fixed at 1000 NFEs). Specifically, we compare our approach against a random selection strategy, as well as fixed step allocations concentrated in the early, middle, and late stages of the denoising trajectory.

As demonstrated in Table~\ref{tab:key_step_selection}, our PCA-based strategy consistently outperforms all heuristic baselines across all metrics. Notably, the \textit{Fixed: Early} strategy yields the poorest overall results, aligning with our main text analysis: early scaling interventions severely disrupt the structure-sensitive generation trajectory. Similarly, the \textit{Fixed: Late} approach proves suboptimal, as late-stage denoising predominantly refines fine-grained visual attributes, yielding marginal scaling benefits. While the \textit{Fixed: Middle} strategy achieves competitive outcomes, our method surpasses it by utilizing PCA-curvature to dynamically pinpoint the effective mid-stage inflection points. This allows us to achieve scaling performance superiority without relying on rigid heuristics.

\begin{table}[t]
\centering
\caption{Performance comparison under equal wall-clock runtime budgets.}
\label{tab:runtime_comparison}
\setlength{\tabcolsep}{10pt}
\renewcommand{\arraystretch}{1.12}
\resizebox{0.9\linewidth}{!}{
\begin{tabular}{llcccc}
\toprule
Runtime & Method & GenEval $\uparrow$ & NIQE $\downarrow$ & CLIP $\uparrow$ & ImageReward $\uparrow$ \\
\midrule
\multirow{4}{*}{50s} 
& BoN & 0.6853 & 4.1483 & 0.2805 & 1.1259 \\
& ZO & 0.6855 & 4.0786 & 0.2764 & 1.1440 \\
& FK & 0.6843 & 4.0933 & 0.2793 & 1.1249 \\
& Ours & \textbf{0.7024} & \textbf{4.0375} & \textbf{0.2842} & \textbf{1.1980} \\
\midrule
\multirow{4}{*}{150s} 
& BoN & 0.7043 & 4.0596 & 0.2885 & 1.2841 \\
& ZO & 0.7122 & 4.0096 & 0.2790 & 1.2986 \\
& FK & 0.7130 & 3.9954 & 0.2853 & 1.2955 \\
& Ours & \textbf{0.7440} & \textbf{3.8792} & \textbf{0.2930} & \textbf{1.3964} \\
\midrule
\multirow{4}{*}{250s} 
& BoN & 0.7157 & 4.0154 & 0.2878 & 1.3479 \\
& ZO & 0.7234 & 3.9486 & 0.2792 & 1.3830 \\
& FK & 0.7216 & 3.9753 & 0.2879 & 1.3687 \\
& Ours & \textbf{0.7583} & \textbf{3.8283} & \textbf{0.2961} & \textbf{1.5111} \\
\bottomrule
\end{tabular}
}
\end{table}

\subsection{Wall-clock/quality Pareto analysis}
\label{Wall-clock}
\textbf{Equal wall-clock time pareto analysis vs. TTS baselines.}
Regarding wall-clock efficiency, we conducted a rigorous equal runtime budget comparison against existing TTS baselines (Best-of-N (BoN), Zero-Order (ZO), and Fliessbach-Kramer (FK)) across three wall-clock constraints (50s, 150s, and 250s) using the same FLUX architecture. As detailed in the Table~\ref{tab:runtime_comparison}, our approach forms a strictly superior Pareto frontier across all time budgets and metrics.

\textbf{Paradigm shift from base sampling and practical Trade-offs.}
While our RTS naturally incurs higher latency than standard TTS-free (single-pass) sampling, this reflects a deliberate paradigm shift rather than a method drawback. Standard sampling operates under a rigid compute budget where generation quality is bound to a fixed ceiling. In contrast, RTS pioneers a controllable trade-off between inference time and generation quality. This paradigm is particularly vital for quality-critical, offline scenarios (e.g., commercial advertising design), where maximizing visual aesthetics vastly outweighs latency constraints.

More importantly, RTS provides a practical solution to the strict GPU VRAM bottlenecks of modern foundation models. Rather than scaling up model parameters, (which demands massive memory) RTS trades offline compute time for generation quality on accessible hardware. As shown in Figure~\ref{fig:sub1} and Table~\ref{fig:sub1}, applying RTS enables a 2B SD v3 model to achieve a GenEval score of 0.7438, remaining highly competitive with the 20B Qwen-Image model (0.7692). Users can thus leverage RTS to match or exceed the performance of 6$\times$ to 10$\times$ larger models without the prohibitive VRAM costs.

\subsection{3D visualization of the reward-guided noise search} 
\begin{figure}[htbp]
\centering
\includegraphics[width=1.0\textwidth]{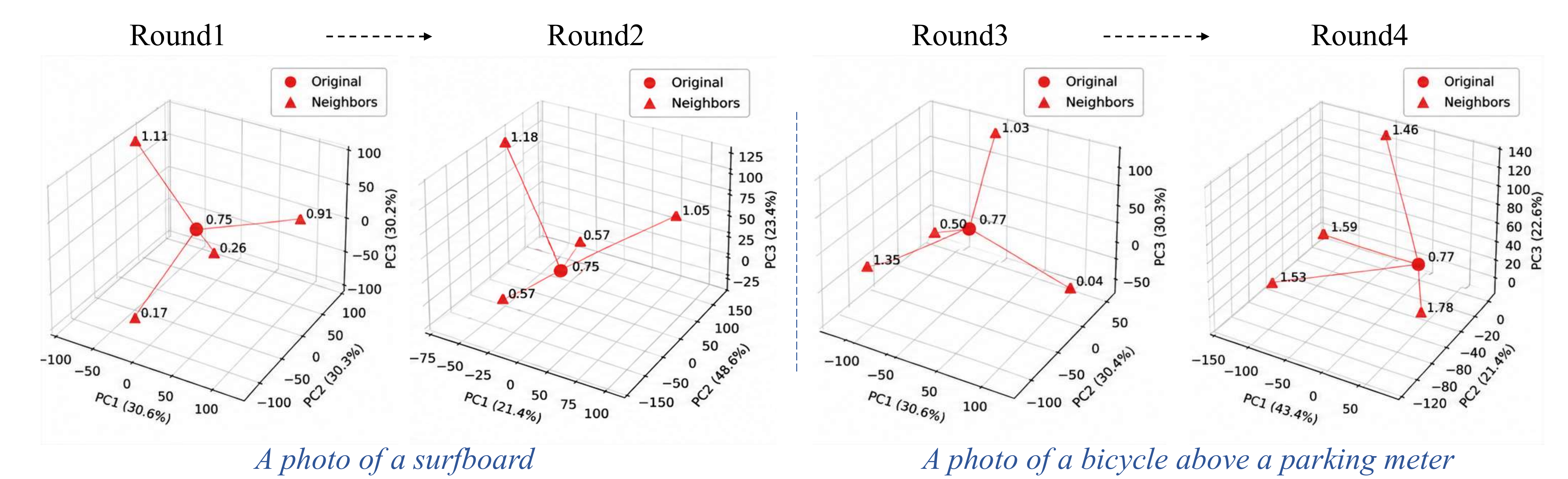} 
\caption{3D PCA visualization of the reward-guided noise search process. Numbers annotated next to each point indicate the corresponding reward value.}
\label{fig_initial1}
\end{figure}

To investigate the underlying optimization dynamics of our framework, we visualize the evolution of noise states in the latent space using Principal Component Analysis (PCA), as shown in Figure~\ref{fig_initial1}. Based on the Flux model, we randomly select two prompts and track the geometric transitions from coarse-grained to fine-grained optimization rounds with the neighbor count set to $N=4$. While the base noise serves as the optimization anchor, the generated neighbor candidates exhibit a distinct directional shift. We observe that the reward scores for these four neighbors improve significantly compared to the initial state. The 3D trajectory clearly indicates that under our Reward-Guided Noise Optimization strategy, the noise distribution collectively migrates towards high-reward regions of the manifold. This geometric evidence confirms that our gradient guidance effectively navigates the latent space, thereby validating the efficacy of our proposed approach.

\section{More visualization results}
\label{sec:more_visual}
\begin{figure}[t]
\centering
\includegraphics[width=0.9\textwidth]{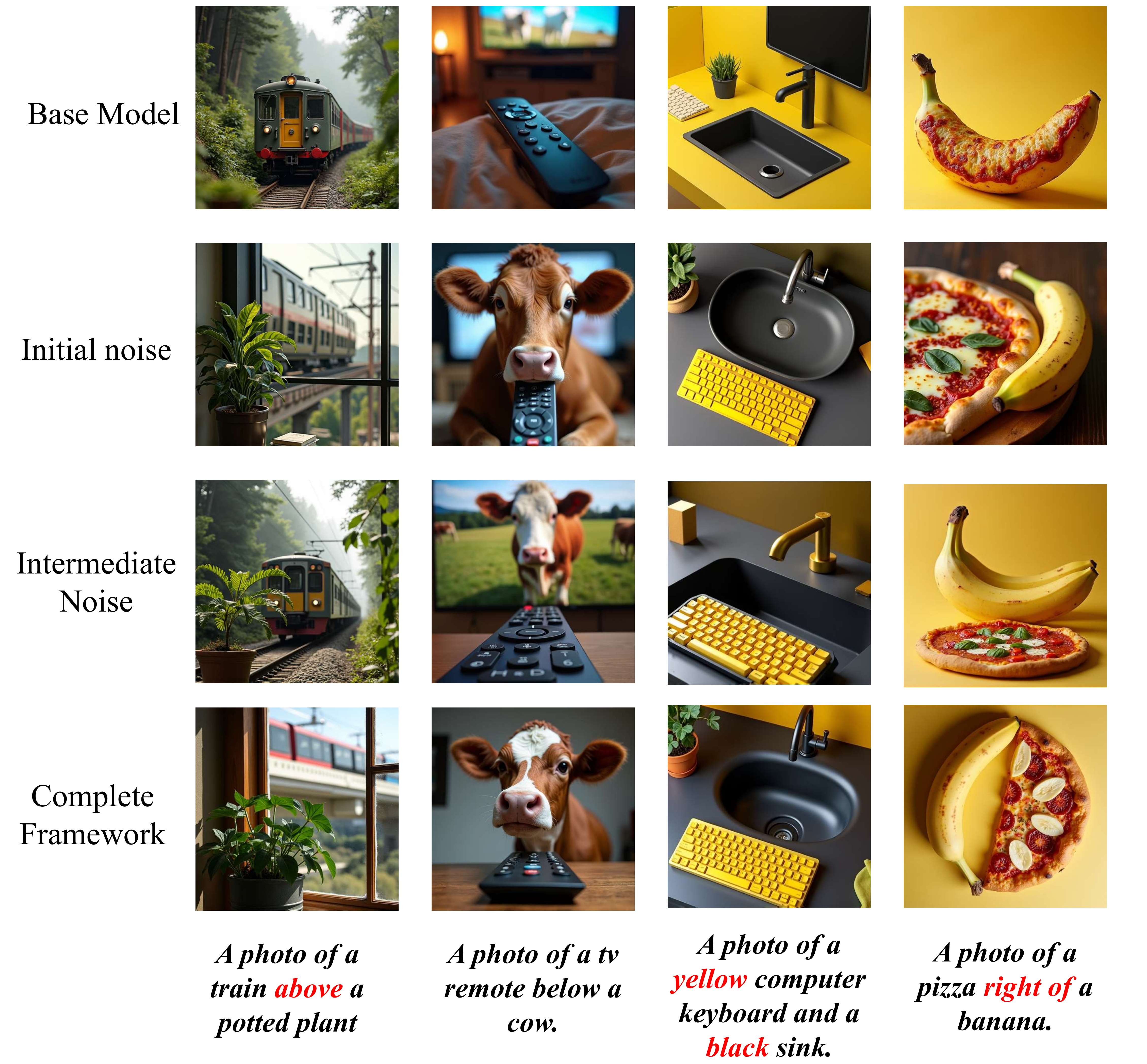} 
\caption{Visualizations of our ablation results on GenEval prompts.}
\label{fig6}
\end{figure}

We conduct visualization experiments on ablation variants of our framework to evaluate the individual contributions of each module. As illustrated in Figure~\ref{fig6}, both the Initial Noise Optimization and Intermediate noise optimization modules improve the semantic consistency of the base model. Specifically, we observe distinct roles for each component: the initial noise optimization induces more explicit, macroscopic changes, significantly influencing image structure and spatial composition. In contrast, the intermediate noise optimization primarily enhances fine-grained detail quality and texture fidelity. However, employing either module in isolation may still result in semantic misalignment or spatial ambiguity, while our complete framework effectively combines their strengths to achieve the best prompt alignment and image fidelity.

\begin{figure}[t]
\centering
\includegraphics[width=1.0\textwidth]{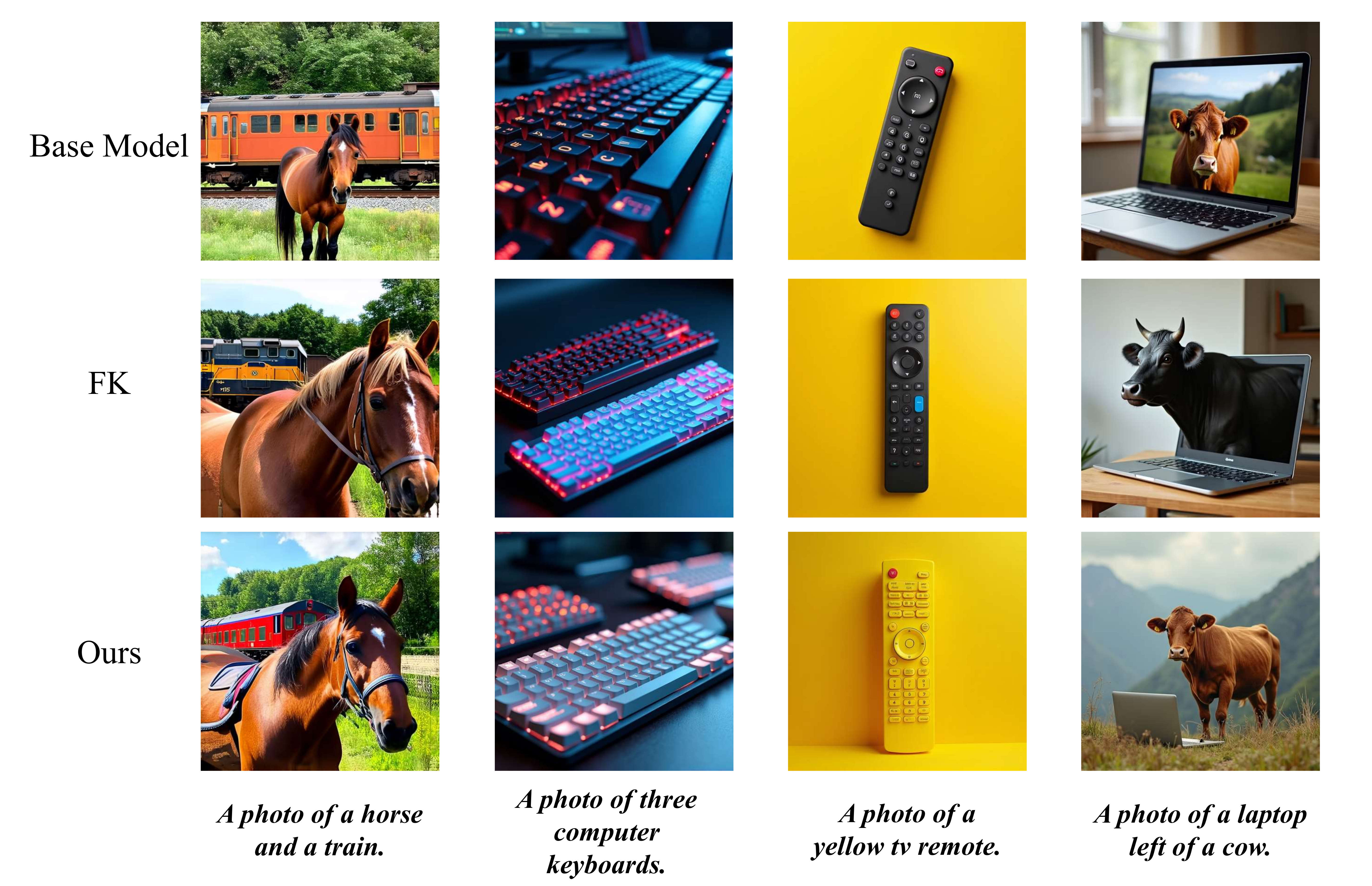} 
\caption{More visualizations of our scaling results compared with the base model and the SOTA method on GenEval prompts. Our approach demonstrates more reliable and stable semantic consistency alongside superior visual fidelity.}
\label{fig7}
\end{figure}

\end{document}